\documentclass[runningheads]{llncs}
\usepackage{graphicx}
\usepackage{comment}
\usepackage{amsmath,amssymb} 
\usepackage{color}
\usepackage{colortbl}

\usepackage{tikz}
\usepackage{comment} 
\usepackage{amsmath,amssymb} 
\usepackage{color}

\usepackage{graphicx}
\usepackage{subfigure}
\usepackage{ulem}
\usepackage{multirow}
\usepackage{makecell}
\usepackage{cite}
\usepackage{caption}
\usepackage{hhline}
\usepackage{bbding}
\usepackage[american]{babel}
\usepackage{microtype}

\begin{document}
\pagestyle{headings}
\mainmatter
\def\ECCVSubNumber{806}  

\title{Blind Face Restoration via Deep Multi-scale Component Dictionaries} 

\definecolor{lightgray}{gray}{.92}
\definecolor{tinygray}{gray}{.96}

\newcommand{\etal}{\textit{et al}.}
\newcommand{\ie}{\textit{i}.\textit{e}.}
\newcommand{\eg}{\textit{e}.\textit{g}.}
\newcommand{\etc}{\textit{etc}}
\newcommand{\para}[1]{\par\noindent\textbf{#1}}

\captionsetup{font={small}}

\titlerunning{Blind Face Restoration via Deep Multi-scale Component Dictionaries}
%
\author{Xiaoming Li\inst{1,4,6}\orcidID{0000\text{-}0003\text{-}3844\text{-}9308} 
	\and
	Chaofeng Chen\inst{2,4}\orcidID{0000\text{-}0001\text{-}6137\text{-}5162} 
	\and 
	Shangchen~Zhou\inst{3}\orcidID{0000\text{-}0001\text{-}8201\text{-}8877} 
	\and
	Xianhui Lin\inst{4}\orcidID{0000\text{-}0002\text{-}8974\text{-}2064} \and \\ 
	Wangmeng Zuo\inst{1,5(}\Envelope$^{)}$\orcidID{0000\text{-}0002\text{-}3330\text{-}783X} 
	\and
	Lei~Zhang\inst{4,6}\orcidID{0000\text{-}0002\text{-}2078\text{-}4215}}
\authorrunning{Xiaoming Li, et al.}
%
\institute{Faculty of Computing, Harbin Institute of Technology, China \and
	Department of Computer Science, The University of Hong Kong \and
	School of Computer Science and Engineering, Nanyang Technological University \and
	DAMO Academy, Alibaba Group \and
	Peng Cheng Lab, Shenzhen \and
	Department of Computing, The Hong Kong Polytechnic University
	\\
	\email{csxmli@hit.edu.cn, cfchen@cs.hku.hk, shangchenzhou@gmail.com, xianhui.lxh@alibaba-inc.com, wmzuo@hit.edu.cn, cslzhang@comp.polyu.edu.hk}
}
\maketitle

\begin{abstract}
Recent reference-based face restoration methods have received considerable attention due to their great capability in recovering high-frequency details on real low-quality images. However, most of these methods require a high-quality reference image of the same identity, making them only applicable in limited scenes. To address this issue, this paper suggests a deep face dictionary network (termed as DFDNet) to guide the restoration process of degraded observations. 
To begin with, we use K-means to generate deep dictionaries for perceptually significant face components (\ie, left/right eyes, nose and mouth) from high-quality images.
Next, with the degraded input, we match and select the most similar component features from their corresponding dictionaries and transfer the high-quality details to the input via the proposed dictionary feature transfer (DFT) block.
In particular, component AdaIN is leveraged to eliminate the style diversity between the input and dictionary features (\eg, illumination),
and a confidence score is proposed to adaptively fuse the dictionary feature to the input.
Finally, multi-scale dictionaries are adopted in a progressive manner to enable the coarse-to-fine restoration.
Experiments show that our proposed method can achieve plausible performance in both quantitative and qualitative evaluation, and more importantly, can generate realistic and promising results on real degraded images without requiring an identity-belonging reference. The source code and models are available at \url{https://github.com/csxmli2016/DFDNet}.
\end{abstract}

\section{Introduction}\label{section1}

Blind face restoration (or face hallucination) aims at recovering realistic details from real low-quality (LQ) image to its high-quality (HQ) one, without knowing the degradation types or parameters. Compared with single image restoration tasks, \eg, image super-resolution~\cite{dong2014learning,wang2018esrgan,zhang2018rcan}, denoising~\cite{zhang2017beyond,zhang2018ffdnet}, and deblurring~\cite{kupyn2018deblurgan, Kupyn_2019_ICCV}, blind image restoration suffers from more challenges, yet is of great practical value in restoring real LQ images. 

Recently, benefited from the carefully designed architecture and the incorporation of related priors in deep neural convolutional networks, the restoration results tend to be more plausible and acceptable. Though great achievements have been made, the real LQ images usually contain complex and diverse distributions that are impractical to synthesize, making the blind restoration problem intractable.
To solve this issue, reference-based methods~\cite{li2018learning,dogan2019exemplar,wang2018recovering,zhang2019image} have been suggested by using reference prior in image restoration task to improve the process of network learning and alleviate the dependency of network on degraded input. 
Among these methods, GFRNet~\cite{li2018learning} and GWAINet~\cite{dogan2019exemplar} adopt a frontal HQ image as reference to guide the restoration of degraded observation. However, these two methods suffer from two drawbacks. 
1) They have to obtain a frontal HQ reference which is from the same identity with LQ image.
2) The differences of poses and expressions between the reference and degraded input will affect the reconstruction performance. 
These two requirements limit their applicative ability to some specific scenarios 
(\eg, old film restoration or phone album that supports identity group). 

In this paper, we present a DFDNet by building deep face dictionaries to address the aforementioned difficulties. 
We note that the four face components (\ie, left/right eyes, nose and mouth) are similar among different people. 
Thus, in this work, we  off-line build face component dictionaries by adopting K-means on large amounts of HQ face images. This manner can obtain more accurate component reference without requiring the corresponding identity-belonging HQ images, 
which makes the proposed model applicable in most face restoration scenes. 
To be specific, we firstly use pre-trained VggFace~\cite{cao2018vggface2} to extract the multi-scale features of HQ face images in different feature scale (\eg, output of different  convolutional layers). Secondly, we adopt RoIAlign~\cite{he2017mask} to crop their component features based on the facial landmarks. K-means is then applied on these features to generate the $K$ clusters for each component on different feature levels. 
After that, component adaptive instance normalization (CAdaIN) is proposed to norm the corresponding dictionary feature which helps to eliminate the effect of style diversity (\ie, illumination or skin color).
Finally, with the degraded input, we match and select the dictionary component clusters which have the smallest feature distance to guide the following restoration process in an adaptive and progressive manner. 
A confidence score is predicted to balance the input component feature and the selected dictionary feature.
In addition, we use multi-scale dictionaries to guide the restoration progressively which further improves the performance.
Compared with the former reference-based methods (\ie, GFRNet~\cite{li2018learning} and GWAINet~\cite{dogan2019exemplar}), which have only one HQ reference, our DFDNet has more component candidates to be selected as a reference, thus making our model achieve superior performance.  

Extensive experiments are conducted to evaluate the performance of our proposed DFDNet. The quantitative and qualitative results show the benefits of deep multi-scale face dictionaries brought in our method. Moreover, DFDNet can also generate plausible and promising results on real LQ images. Without requiring identity-belonging HQ reference, our method is flexible and practical in most face restoration applications. 

To sum up, the main contributions of this work are:
\begin{itemize}
	\item We use deep component dictionaries as reference candidates to guide the degraded face restoration. The proposed DFDNet can generalize to face images without requiring the identity-belonging HQ reference, which is more applicative and efficient than those reference-based methods. 
	%
	\item We suggest a DFT block by utilizing CAdaIN to eliminate the distribution diversity between the input and dictionary clusters for better dictionary feature transfer, and we also propose a confidence score to adaptively fuse the dictionary feature to the input with different degradation level.
	\item We adopt a progressive manner for training DFDNet by incorporating the component dictionaries in different feature scales. This can make our DFDNet learn coarse-to-fine details. 
	\item Our proposed DFDNet can achieve promising performance on both synthetic and real degraded images, showing its potential in real applications.
\end{itemize}

\section{Related Work}\label{section2}
In this section, we discuss recent works about single image and reference-based image restoration methods which are closely related to our work.
\subsection{Single Image Restoration}
Along with the benefits brought by deep CNNs, single image restoration has achieved great success in many tasks, \eg, image super-resolution~\cite{dong2014learning,kim2016accurate,ledig2017photo,zhang2018rcan,Zhang_2019_CVPR}, denoising~\cite{zhang2017beyond,zhang2018ffdnet,yang2017bm3d,guo2019toward}, deblurring~\cite{nah2017deep,kupyn2018deblurgan,zhang2019deep}, and compression artifact removal~\cite{dong2015compression,galteri2017deep,guo2017one}. Due to the specific facial structure, there are also several well-developed methods for face hallucination~\cite{zhu2016deep,cao2017attention,huang2017wavelet,xu2017learning,chrysos2017deep,Yu_2018_ECCV,Yu_2018_CVPR,progressive_face_sr,Chen_2018_CVPR}. Among these methods, Huang \etal~\cite{huang2017wavelet} suggest to  ultra-resolve a very low resolution face image by using the neural networks to predict the wavelet coefficients of HQ images. Cao \etal~\cite{cao2017attention} propose reinforcement learning to discover the attended regions and then enhance them with a learnable local network. To better recover the structure details, there are also some methods that incorporate the image prior knowledge in the restoring process. Wang \etal~\cite{wang2018recovering} propose to use semantic segmentation probability maps as class prior to recover class-aware textures on natural image super-resolution task. It firstly takes the LR images through a segmentation network to generate the class probability maps. And then these maps and LQ features are fused together by spatial feature transformation. As for face images, Shen \etal~\cite{shen2018deep} propose to learn a global semantic face prior as input to impose local structure on the output. Similarly, Xu \etal~\cite{Yu_2018_ECCV} use a multi-tasks model to predict the facial components heatmaps and use them for incorporating structure information. Chen \etal~\cite{Chen_2018_CVPR} learn the facial geometry prior (\ie, landmarks heatmaps and parsing maps) and take them to recover the high-resolution results. Yu \etal~\cite{Yu_2018_CVPR} develop a facial attribute-embedded network by incorporating face attributes vector in the LR feature space. Kim \etal~\cite{progressive_face_sr} adopt a progressive manner to generate the successive higher resolution output and propose a facial attention loss on landmarks to constrain the structure of reconstruction. However, most of these facial prior knowledge mainly focus on geometry constrains (\ie, landmarks or heatmaps), which may not bring direct facial details for the restoration of LQ image. Thus, most of these single image restoration methods failed to generate plausible and realistic details on real LQ face images because of the ill-posed problem and the limitation of a single image or facial structure prior brought to the learning process of networks.
\subsection{Reference-Based Image Restoration}
Due to the limitation of single image restoration methods on real-world LQ images, there are some works that use an additional image to guide the restoration process, which can bring the object structure details to the final result.  As for natural image restoration, Zhang \etal~\cite{zhang2019image} utilize a reference image which has similar content with a LR image and then adopt a global matching scheme to search the similar content patches. These reference feature patches are then used to swap the texture feature of LR images. This method can achieve great visual improvements. However, it is very time and memory consuming in searching similar patches from the global content. Moreover, the requirement of reference further limits its application, because finding a natural image with a similar content for each LR input is also terrible and sometimes it is impossible to obtain these types of image.  

Different from natural image, face owns specific structures and share the similar components on different images of the same identity. Based on this observation, two reference-based methods have been developed for face restoration. Li \etal~\cite{li2018learning} and Dogan \etal~\cite{dogan2019exemplar} use a fixed frontal HQ reference for each identity to provide identity-aware features to benefit the restoration process. However, we note that face images are usually taken under unconstrained conditions, \eg, different background, poses, expressions, illuminations, \etc. To solve this problem, they utilize a WarpNet to predict flow field to warp the reference to align with the LQ image. However, the alignment still does not solve all the differences between the reference and input, \ie, mouth close to open. Besides, the warped reference is usually unnatural and may take obvious artifacts to the final reconstruction result. We note that each component between different identity has the similar structure (\ie, teeth, nose, and eyes). It is intuitive to split the whole face into different parts, and generate the representative components for each one. To achieve this goal, we firstly use K-means on HQ images to cluster different component features off-line. Then we match the LQ features from the conducted component dictionaries to select the one with the similar structures to guide the latter restoration. Moreover, with the conducted dictionaries, we do not require an identity-belonging reference anymore, and more component candidates can be selected as reference. It is much more accurate and effective than only one face image in reference-based restoration and can be applied in the unconstrained applications.

\section{Proposed Method}\label{section3}

\begin{figure*}[!t]
	\centering
	\subfigure[Off-line generation of multi-scale component dictionaries.]{
		\begin{minipage}{1\textwidth}
			\centering
			\includegraphics[width=0.95\textwidth]{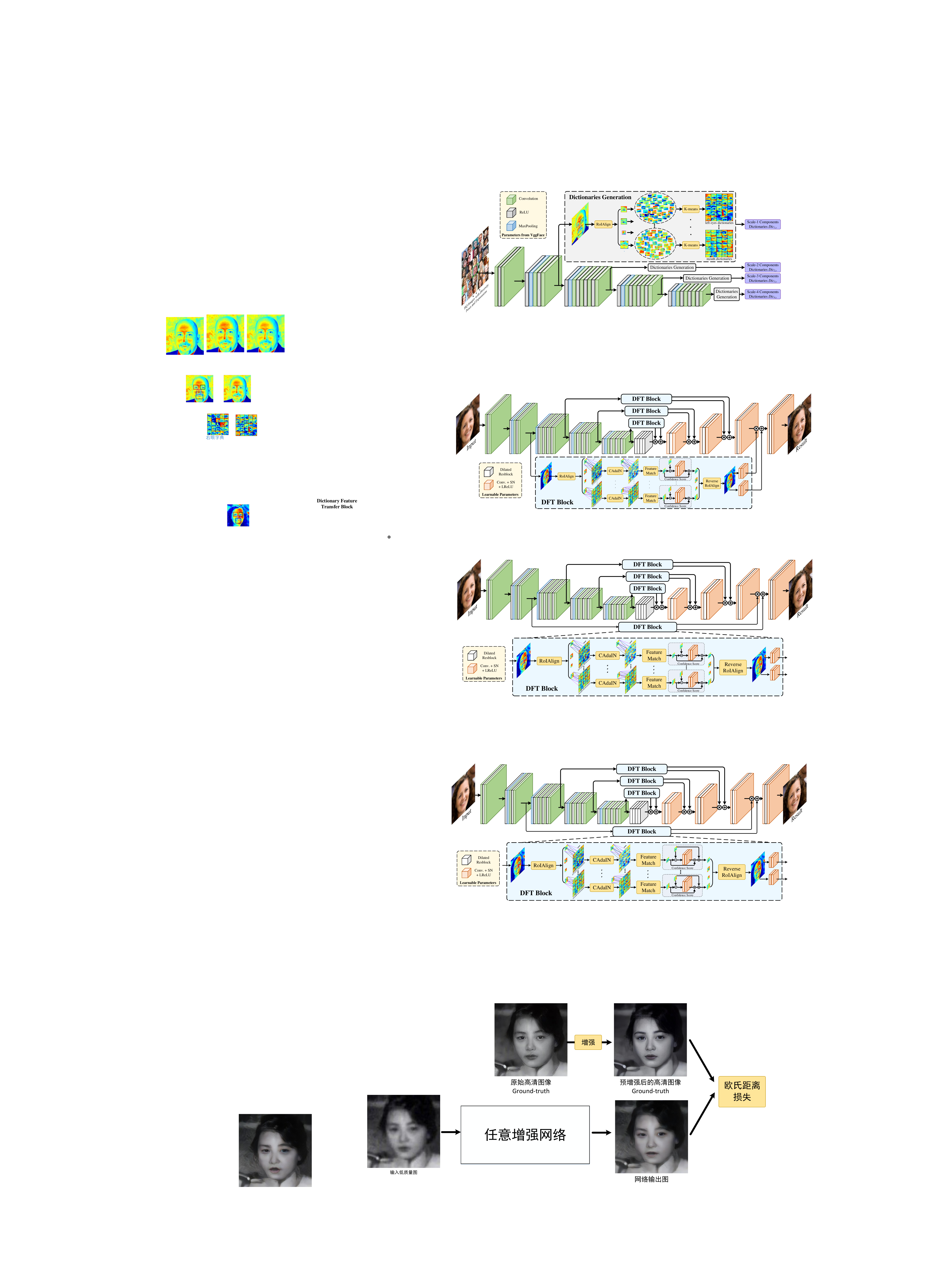}
		\end{minipage}
	}
	\subfigure[Architecture of our DFDNet for dictionary feature transfer.]{
		\begin{minipage}{1\textwidth}
			\centering
			\includegraphics[width=0.95\textwidth]{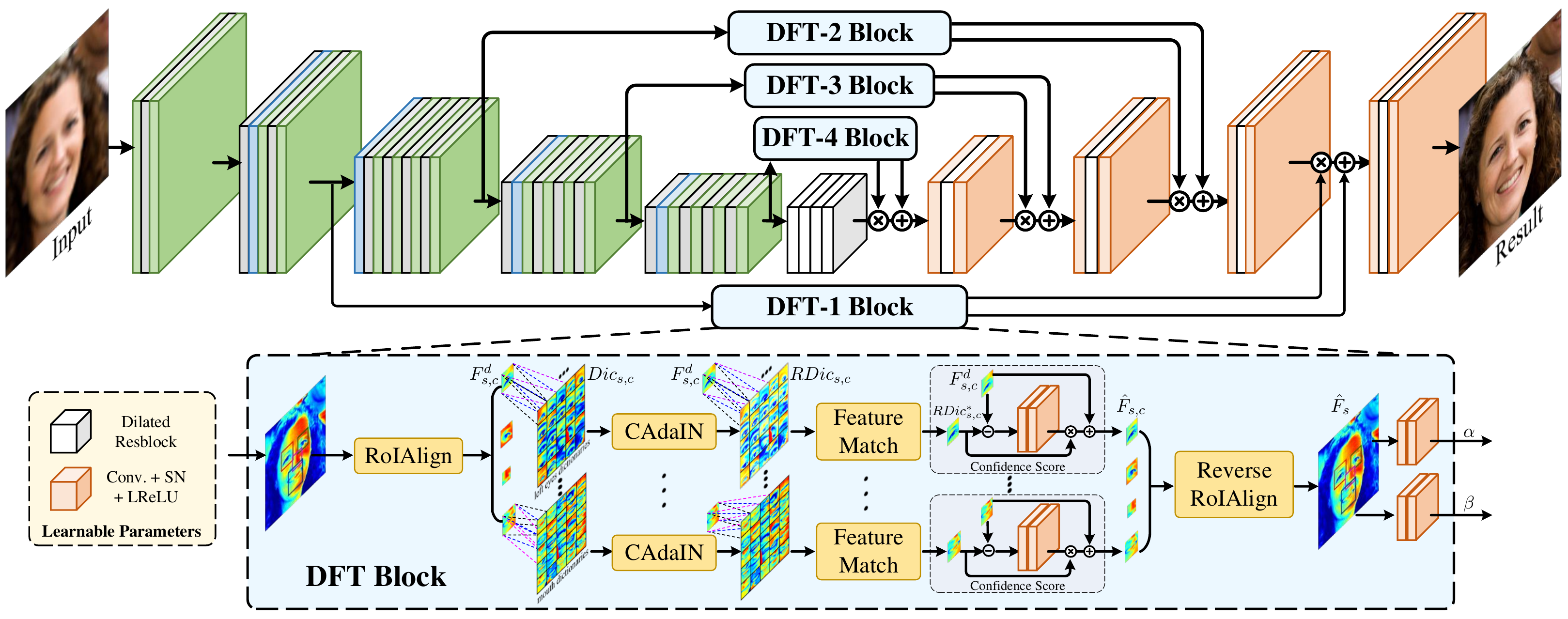}
		\end{minipage}
	}
	\caption{Overview of our proposed method. It mainly contains two parts: (a) the off-line generation of multi-scale component dictionaries from large amounts of high-quality images which have diverse poses and expressions. K-means is adopted to generate $K$ clusters for each component (\ie, left/right eyes, nose and mouth) on different feature scales.
		(b) The restoration process and dictionary feature transfer (DFT) block that are utilized to provide the reference details in a progressive manner. Here, DFT-$i$ block takes the Scale-$i$ component dictionaries for reference in the same feature level.}
	\label{fig:pipeline}
\end{figure*}
Inspired by the former reference-based image restoration methods~\cite{zhang2019image, li2018learning, dogan2019exemplar}, this work attempts to overcome the limitation of requiring reference image in face restoration. Given a LQ image $I^d$, our proposed DFDNet aims to generate plausible and realistic HQ one $\hat{I}^{h}$ with the conducted component dictionaries. 
The whole pipeline is shown in Fig.~\ref{fig:pipeline}. In the first stage (Fig.~\ref{fig:pipeline} (a)), we firstly generate the deep component dictionaries from the high-quality images $I^{h}$ via k-means. These dictionaries can be selected as candidate component references. In the second stage (Fig.~\ref{fig:pipeline} (b)), for each component of the degraded observation $I^d$, 
our DFDNet selects the dictionary features that have the most similar structure with the input. 
Specially, we re-norm the whole dictionaries via component AdaIN (termed as CAdaIN) based on the input component to eliminate the distribution or style diversity. The selected dictionary features are then utilized to guide the restoration process via dictionary feature transformation. Furthermore, we introduce a confidence score on the selected dictionary feature to generalize different degradation levels through weighted feature fusion. The progressive manner from coarse to fine is also beneficial to the restoration process. 
In the following, we first describe the off-line generation of multi-scale deep component dictionaries. Then the details of our proposed DFDNet along with the dictionaries feature transfer (DFT) blocks are interpreted. The objective functions for training are finally presented.

\setcounter{footnote}{0}
\subsection{Off-line Generation of Component Dictionaries}\label{dictionary}
To build the deep component dictionaries that cover the most types of faces, we adopt FFHQ dataset~\cite{karras2019style} due to its high-quality and considerable variation in terms of age, ethnicity, pose, expression, \etc. We utilize DeepPose~\cite{Ruiz_2018_CVPR_Workshops} and Face$^\text{++}$\footnote{https://www.faceplusplus.com.cn/emotion-recognition/} to recognize their poses and expressions (\ie, anger, disgust, fear, happiness, neutral, sadness and surprise), respectively, to balance the distribution of each attribute. Among these 70,000 high-quality images of FFHQ, we select 10,000 ones to build our dictionaries. Given a high-quality image $I^h$, we first use pre-trained VggFace~\cite{cao2018vggface2} to extract its features on different scales. With the facial landmarks $L^h$ detected by dlib~\cite{dlib09},
we utilize RoIAlign~\cite{he2017mask} 
to crop and re-sample these four components on each scale to a fixed size. We then adopt K-means~\cite{scikit_learn} to generate $K$ clusters for each component, resulting in our component dictionaries. In particular, for handling $256\times256$ images, the feature sizes of left/right eyes, nose and mouth on scale-1 are set to 40/40, 25, 55, respectively. The sizes are down-sampled one by one by two times for the following scale-\{2, 3, 4\}. These dictionary feature can be formulated as:
\begin{equation}
\small
\label{eqn:kmeans}
Dic_{s,c} = \mathcal{F}_{Dic}\left(I^{h} | L^{h}; \Theta_{Vgg}\right),
\end{equation}
where $s\in \{1,2,3,4\}$ is the dictionary scale, $c\in\{\text{left eye}, \text{right eye}, \text{nose}, \text{mouth}\}$ is the type of components, and $\Theta_{Vgg}$ is the fixed parameters from VggFace.

\subsection{Deep Face Dictionary Network}
After building the high-quality component dictionaries, our DFDNet is then proposed to transfer the dictionary features to the degraded input $I^d$. The proposed DFDNet can be formulated as:
\begin{equation}
\small
\hat{I} = \mathcal{F}(I^d|L^d, Dic; \Theta),
\end{equation}
where $L^d$ and $Dic$ represent the facial landmarks of $I^d$ and the component dictionaries in Eqn.~\ref{eqn:kmeans}, respectively. $\Theta$ denotes the learnable parameters of DFDNet.

To guarantee the features of $I^d$ and $Dic$ in the same feature space, we take the pre-trained VggFace model as the encoder of DFDNet, which has the same network architecture and parameters in the dictionary generation network (Fig.~\ref{fig:pipeline} (a)). Suppose that the encoder of DFDNet is different from VggFace or trainable in the training phase, it easily generates different features which are inconsistent with the pre-conducted dictionaries.
For better transferring the dictionary feature to the input components, we suggest a DFT block and use it in a progressive manner. It mainly contains five parts, \ie, RoIAlign, CAdaIN, Feature Match, Confidence Score and Reverse RoIAlign. As for the encoder features of $I^d$, we first utilize RoIAlign to generate four component regions. We note that these input components may have different distribution/style with the cluster of conducted dictionaries $Dic_{s,c}$, we here suggest a component adaptive instance norm~\cite{huang2017arbitrary} (CAdaIN) to re-norm each cluster in the dictionaries. The feature match scheme is then utilized to select the cluster with the similar texture. In addition, a confidence score is predicted based on the residual between the selected cluster and the input feature to better provide complementary details on input. The reverse RoIAlign is finally adopted to paste the restored features to the corresponding locations. For better transformation of restored features to the decoder, we modify the UNet~\cite{ronneberger2015u} and propose to use spatial feature transform (SFT)~\cite{wang2018recovering} to transfer the dictionary features to the degraded input.
\subsubsection{CAdaIN.} 
We note that face images are usually under unconstrained conditions, \eg, different illuminations, skin color. To eliminate the effect of these diversities between the input components and dictionaries, we adopt component AdaIN (CAdaIN) to re-norm the clusters in component dictionaries for accurate feature matching. AdaIN~\cite{huang2017arbitrary} can remain the structure while translate the content to the desired style. Denote $F^{d}_{s,c}$ and $Dic_{s,c}^{k}$ as the $c\text{-}th$ component features of the input $I^d$ and the $k\text{-}th$ cluster from the component dictionaries at scale $s$, respectively. The re-normed dictionaries $RDic_{s,c}$ by CAdaIN is formulated by:
\begin{equation}
\small
\label{eqn:cadain}
RDic_{s,c}^{k}=\sigma\left(F^{d}_{s,c}\right)\left(\frac{Dic_{s,c}^{k}-\mu\left(Dic^{k}_{s,c}\right)}{\sigma\left(Dic_{s,c}^{k}\right)}\right)+\mu\left(F^{d}_{s,c}\right)
\end{equation}
where $s$ and $c$ are the dictionary scale and the type of components defined in Eqn.~\ref{eqn:kmeans}. $\sigma$ and $\mu$ are the mean and standard deviation. The re-normed dictionaries $RDic_{s,c}^{k}$ has the similar distribution with input components $F_{s,c}^d$, 
which can not only eliminate the style difference, but also facilitate the feature match scheme.
\subsubsection{Feature Match.} As for the input component feature $F^{d}_{s,c}$ and the re-normed dictionaries $RDic_{s,c}$, we adopt inner product to measure the similarity between the $F^{d}_{s,c}$ and all the clusters in $RDic_{s,c}$. For $k\text{-}th$ cluster in component dictionary, the similarity is defined as:
\begin{equation}
\small
S^{k}_{s,c}=\left\langle F^{d}_{s,c}, RDic_{s,c}^{k}\right\rangle,
\end{equation}
The input component feature $F^{d}_{s,c}$ matches across all the clusters in the re-normed component dictionaries to select the most similar one. $F^{d}_{s,c}$ has the same size with $k\text{-}th$ cluster in the corresponding dictionaries, thus this inner product operation can be regarded as a convolutional layer with zero bias and weights of $F^{c}_{s,d}$ performed over all the clusters. 
This is very efficient to obtain the dictionaries' similarity scores. Among all the scores $S_{s,c}$, we select the re-normed cluster with the highest similarity as the matched dictionaries, termed as $RDic^{*}_{s,c}$. This selected component feature $RDic^{*}_{s,c}$ is then utilized to provide the high-quality details to guide the restoration of the input components in the following section.

\subsubsection{Confidence Score.} We note that the slight degradation of input (\eg, $\times2$ super-resolution) relies little on the dictionaries and vice versa. To generalize our DFDNet to different degradation level, we take the residual between $F^{d}_{s,c}$ and $RDic^{*}_{s,c}$ as input to predict a confidence score that performs on the selected dictionary feature $RDic^{*}_{s,c}$. The result is expected to contain the absent high-quality details which can add back to $F^{d}_{s,c}$. The output of confidence score can be formulated by:
\begin{equation}
\small
\hat{F}_{s,c}= F^{d}_{s,c} + RDic^*_{s,c} *  \mathcal{F}_{Conf}(RDic^{*}_{s,c} - F^{d}_{s,c};\Theta_{C}),
\end{equation}
where $\Theta_{C}$ is the learnable parameters of confidence score block $\mathcal{F}_{Conf}$. 

\subsubsection{Reverse RoIAlign.} After all the input components are processed by the former section, here we utilize a reverse operation of RoIAlign by taking $\hat{F}_{s,c}$ and $c \in \{ {\text{left/right eyes, nose and mouth}}\}$ to their original locations of $F^{d}_{s,c}$. Denote the result of reverse RoIAlign $\hat{F}_{s}$. This manner can easily keep and translate other features (\eg, background)  to the decoder for better restoration.

Inspired by SFT~\cite{wang2018recovering}, which is proposed to learn a feature modulation function that incorporates some prior condition through affine transformation. The scale $\alpha$ and shift $\beta$ parameters are learned from the restored features $\hat{F}_s$ with two convolutional layers. The scale-$s$ SFT layer is formulated as:
\begin{equation}
\small
SFT_s = \alpha \odot F_s^{decoder} + \beta\,,
\end{equation}
where $\alpha$ and $\beta$ are both element-wise weights which have the same shape (i.e., height, width, number of channels) with $F_s^{decoder}$.
After the progressive DFT block, our DFDNet can gradually learn the fine details for the final result $\hat{I}$.

\subsection{Model Objective}
The learning objective for training our DFDNet contains two parts, 1) reconstruction loss that constrains the result $\hat{I}$ close to the ground-truth $I^{h}$, 2) adversarial loss~\cite{goodfellow2014generative} for recovering realistic details.

\subsubsection{Reconstruction Loss.} We adopt mean square error (MSE) on both pixel and feature space (perceptual loss~\cite{johnson2016perceptual}). The whole reconstruction loss is defined as,
\begin{equation}
\small
\label{eqn:perceptual}
\mathcal{L}_{rec} = \lambda_{l2}\|\hat{I}-I^h\|^{2} + \sum_{m=1}^{M}\frac{\lambda_{p,m}}{C_{m} H_{m} W_{m}}\left\|\Psi_{m}(\hat{I})-\Psi_{m}(I^h)\right\|^{2}
\end{equation} 
where $\Psi_{m}$ denotes the $m\text{-}th$ convolution layer of VggFace model $\Psi$. $C$, $H$ and $W$ are the channel, height, and width for the $m\text{-}th$ feature. $\lambda_{l2}$ and $\lambda_{p,m}$ are the trade-off parameters. The first term tends to generate blurry results, while the second one (perceptual loss) is beneficial for improving visual quality for the reconstruction results. The combination of the two terms is common in computer vision tasks and also is effective in the stable training of neural networks. In our experimental settings, we set $M$ equal to 4.

\subsubsection{Adversarial Loss.} It is widely used to generate realistic details in image restoration tasks. In this work, we adopt multi-scale discriminators~\cite{wang2018pix2pixHD} at different size of the restoration results. Moreover, for stable training of each discriminator, we adopt SNGAN~\cite{miyato2018spectral} by incorporating the spectral normalization after each convolution layer. The objective function for training multi-scale discriminators is defined as:
\begin{equation}
\small
\ell_{\mathrm{adv}, D_r}= \sum_{r}^{R} \mathbb{E}_{I^{h}_{\downarrow r} \sim P(I^{h}_{\downarrow r})}\left[\min \left(0,D_r(I^{h}_{\downarrow r})-1\right)\right] 
+\mathbb{E}_{\hat{I}_{\downarrow r} \sim P\left(\hat{I}_{\downarrow r}\right)}\left[\min \left(0,-1-D_{r}(\hat{I}_{\downarrow r})\right)\right],
\end{equation}
where $_{\downarrow r}$ denotes the down-sampling operation with scale factor $r$ and $r \in \{1,2,4,8\}$. Similarly, the loss for training generator $\mathcal{F}$ is defined as:
\begin{equation}
\small
\ell_{\mathrm{adv}, G}=-\lambda_{a,r}\sum_{r}^{R}\mathbb{E}_{I^{d} \sim P\left(I^{d}\right)}\left[D_r\left(\mathcal{F}\left(I^{d} | L^{d}, Dic ; \Theta\right)_{\downarrow r}\right)\right],
\end{equation}
where $\lambda_{a,r}$ is the trade-off parameters for each scale discriminator. 

To sum up, the full objective function for training our DFDNet can be written as the combination of reconstruction and adversarial loss,
\begin{equation}
\mathcal{L} = \ell_{rec} + \ell_{\mathrm{adv}, G}.
\end{equation}

\section{Experiments}
Since the performance of reference-based methods are usually superior to other single image or face restoration methods~\cite{li2018learning}, in this paper, we mainly compare our DFDNet with reference-based (\ie, GFRNet~\cite{li2018learning}, GWAINet~\cite{dogan2019exemplar}) and face prior-based methods (\ie, Shen \etal~\cite{shen2018deep}, Kim \etal~\cite{progressive_face_sr}). 
We also report the results of single natural image (\ie, RCAN~\cite{zhang2018rcan}, ESRGAN~\cite{wang2018esrgan}) and face (\ie, WaveletSR~\cite{huang2017wavelet}) super-resolution methods.
Among these methods, Shen \etal~\cite{shen2018deep} and Kim \etal~\cite{progressive_face_sr} can only handle $128\times128$ images,
while others can restore $256\times256$ images. For fair comparisons, our DFDNet is trained on these two sizes (termed as DFDNet128 and DFDNet256). RCAN~\cite{zhang2018rcan} and ESRGAN~\cite{wang2018esrgan} were originally trained on the natural images, thus we retrain them using our training data for further fair comparison (termed as *RCAN and *ESRGAN). WaveletSR~\cite{huang2017wavelet} was also retrained by using our training data with their released training code (termed as *WaveletSR). Following \cite{li2018learning}, PSNR, SSIM and LPIPS~\cite{zhang2018perceptual} are reported on the super-resolution task ($\times4$ and $\times8$) which also has the random injection of Gaussian noise and blur operation for quantitatively evaluating on the blind restoration task. In terms of qualitative comparison, we demonstrate the comparisons on the synthetic and real-world low-quality images. More visual results including high resolution restoration performance (\ie, $512\times512$) can be found in our supplemental materials. 

\subsection{Training Details}
As mentioned in Section~\ref{dictionary}, we select 10,000 images from FFHQ~\cite{karras2019style} to build our component dictionaries. We note that GFRNet, GWAINet and WaveletSR adopt VggFace2~\cite{cao2018vggface2} as their training data, we also use it for training and validating our DFDNet for fair comparison. To evaluate the generality of our method, we build two test datasets, \ie, 2,000 test images from VggFace2~\cite{cao2018vggface2} which are not overlapped with the training data, and another 2,000 images from CelebA~\cite{liu2015faceattributes}. Each of them has a high-quality reference from the same identity for running GFRNet and GWAINet. 
To synthesize the training data that approximate to the real LQ images, we adopt the same degradation model suggested in GFRNet~\cite{li2018learning},
\begin{equation}
I^{d}=\left((I^h \otimes \mathbf{k}) _{\downarrow_{r}}+\mathbf{n}_{\sigma}\right)_{JPEG_{q}}
\end{equation}
where $\mathbf{k}$ denotes two common types of blur kernel, \ie, Gaussian blur with $\varrho \in \{1:0.1:5\}$ and 32 motion blur kernels from~\cite{levin2009understanding,boracchi2012modeling}. Down-sampler $r$, Gaussian noise $n_{\sigma}$ and JPEG compression quality $q$ are randomly sampled from $\{1:0.1:8\}$, $\{0:1:15\}$ and $\{40:1:80\}$, respectively.
The trade-off parameters for training DFDNet are set as follows: $\lambda_{l2}=100$, $\lambda_{p,1}=0.5$, $\lambda_{p,2}=1$, $\lambda_{p,3}=2$, $\lambda_{p,4}=4$, $\lambda_{a,1}=4$, $\lambda_{a,2}=2$, $\lambda_{a,4}=1$, $\lambda_{a,8}=1$.
The Adam optimizer~\cite{kingma2014adam} is adopted to train our DFDNet with learning rate $lr=2\times10^{-4}$, $\beta_1=0.5$ and $\beta_2=0.999$. $lr$ is reduced by 2 times when the reconstruction loss on validation set becomes non-decreasing. 
The whole model including the generation of multi-scale component dictionaries and the training of DFDNet are executed on a server with 128G RAM and 4 Tesla V100. It takes 4 days to train our DFDNet.

\subsection{Results on Synthetic Images}
\subsubsection{Qualitative evaluation.} The quantitative results of these competing methods on super-resolution task are shown in Table~\ref{tab:quan}. We can have the following observations: 1) Compared with all the competing methods, our DFDNet is superior to others by a large margin on two datasets and two super-resolution tasks (\ie, at least 0.4 dB in $\times 4$ and 0.3 dB in $\times 8$ higher than the 2-\textit{nd} best method). 2) Even though the retrained *RCAN and *ESRGAN have achieved great improvements, the performance is still inferior to GFRNet, GWAINet and our DFDNet, mainly due to the lack of high-quality facial references. 3) With the same training data, reference-based methods (\ie, GFRNet~\cite{li2018learning} and GWAINet~\cite{dogan2019exemplar}) outperform other methods, but are still inferior to our DFDNet, which can be attributed to the incorporation of high-quality component dictionaries and the progressive dictionary feature transfer manner. Given a LQ image, our DFDNet has more candidates to be selected as component reference, resulting in the flexible and effective restoration. 4) Our component dictionaries are conducted on FFHQ~\cite{karras2019style} and DFDNet is trained on VggFace2~\cite{cao2018vggface2}, but the performance on CelebA~\cite{liu2015faceattributes} still outperforms other methods, indicating the great generalization of our DFDNet.
\begin{table}[t]
	\centering
	\scriptsize
	\renewcommand\arraystretch{1.3}
	\caption{{Quantitative comparisons on two datasets and two tasks ($\times$4 and $\times$8).}}
	\setlength{\tabcolsep}{0.5mm}
	{
		
		\begin{tabular}{|c| c c c | c c c | c c c | c c c |}
			\hline
			\rowcolor{lightgray}
			& \multicolumn{6}{c|}{\textbf{VggFace2~\cite{cao2018vggface2}}} & \multicolumn{6}{c|}{\textbf{CelebA~\cite{liu2015faceattributes}}}\\
			\hhline{>{\arrayrulecolor{lightgray}}-|>{\arrayrulecolor{black}}------------}
			\rowcolor{lightgray}& \multicolumn{3}{c|}{$\times 4$} & \multicolumn{3}{c|}{$\times 8$} & \multicolumn{3}{c|}{$\times 4$} & \multicolumn{3}{c|}{$\times 8$} \\
			\rowcolor{lightgray}\multirow{-3}{*}{\makecell[c]{\textbf{Methods}}}&
			\tiny PSNR$\uparrow$ &\tiny SSIM$\uparrow$ & \tiny LPIPS$\downarrow$ & \tiny PSNR$\uparrow$ & \tiny SSIM$\uparrow$ & \tiny LPIPS$\downarrow$ & \tiny PSNR$\uparrow$ & \tiny SSIM$\uparrow$ &  \multicolumn{1}{c|}{\tiny LPIPS$\downarrow$} & \tiny PSNR$\uparrow$ & \tiny SSIM$\uparrow$ & \tiny LPIPS$\downarrow$\\
			\hline \hline
			Shen \etal~\cite{shen2018deep}        & 20.56  & .745   & .080    &   18.79    & .717  & .126  & 21.04 & .751 & .079 & 18.64 & .714 & .131   \\
			Kim \etal~\cite{progressive_face_sr}        & -  & - & -  &   20.99    &   .759  & .095  & - & - & - & 20.72 & .749 & .104   \\
			DFDNet128  & 25.76  & .893   &  .035   &  23.42 &  .841   & .071 & 25.92 & .899 & .031 & 23.40 & .839 & .080    \\
			\hline
			RCAN~\cite{zhang2018rcan} & 24.87  & .889   & .283    &  21.36   & .819  & .295  & 24.93 & .892 & .267 & 21.11 & .814 & .302 \\
			*RCAN  & 25.32  & .896  & .247   & 22.94 &  .836   & .271    & 25.47 & .901 & .217 & 22.84 & .831 & .283 \\
			ESRGAN~\cite{wang2018esrgan} & 24.13  & .876 & .223  & - & - & - & 24.31 & .878 & .210 & - & - & - \\
			*ESRGAN & 24.91  & .891 & .194 & - & - & - & 25.04 & .896 & .193 & - & - & - \\
			WaveletSR~\cite{huang2017wavelet} & 24.30 & .878 & .236 & 21.70 & .823 & .273  & 24.51 & .884 & .247 & 21.42 & .820 & .279 \\
			GFRNet~\cite{li2018learning} &  27.13 & .912   &   .132   &    23.37     &  .856   &  .269 & 27.32 & .915 & .124 & 23.12 & .852 & .273 \\
			GWAINet~\cite{dogan2019exemplar} & - & - & - & 23.41 & .860 & .260 & - & - & - & 23.38 & .859 & .270 \\
			\textbf{DFDNet256} & \textbf{27.54} & \textbf{.923} & \textbf{.114}   & \textbf{23.73} & \textbf{.872} & \textbf{.239} & \textbf{27.77} & \textbf{.925} & \textbf{.103} & \textbf{23.69} & \textbf{.872} & \textbf{.241} \\
			\hline
	\end{tabular}}
	
	\label{tab:quan}
\end{table}

\begin{figure*}[t]
	\setlength{\abovecaptionskip}{3pt}
	\setlength{\belowcaptionskip}{-16pt}
	\centering
	\includegraphics[width=1.01\textwidth]{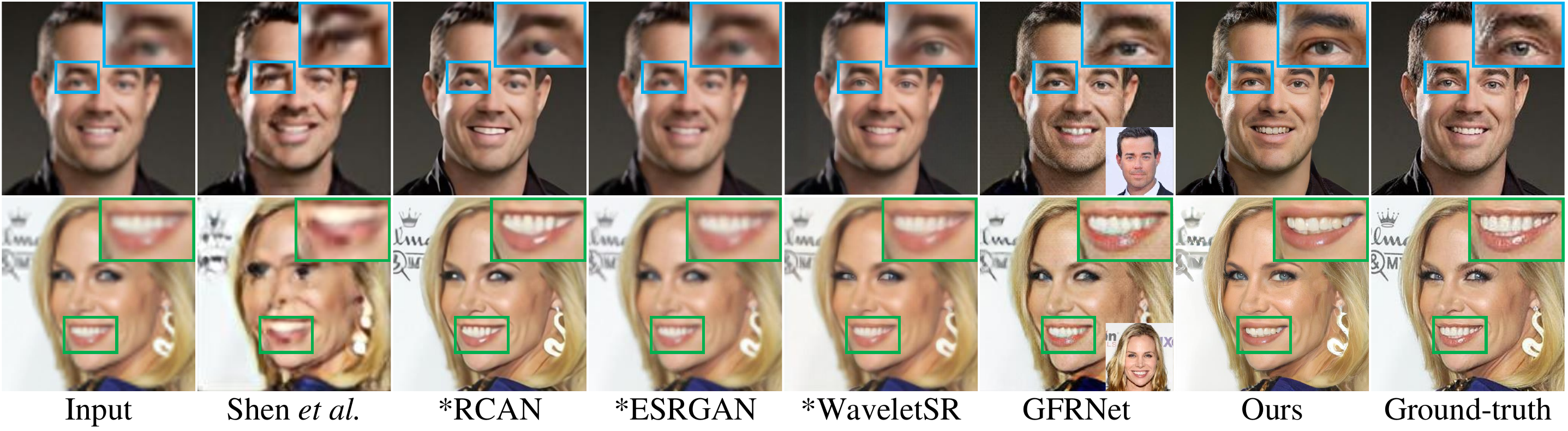}
	\caption{{Visual comparisons of these competing methods on $\times4$ SR task. Close-up in the right bottom of GFRNet is the required guidance.}}
	\label{fig:x4}
\end{figure*}

\begin{figure*}[t]
	\setlength{\abovecaptionskip}{3pt}
	\setlength{\belowcaptionskip}{-6pt}
	\centering
	\includegraphics[width=1.01\textwidth]{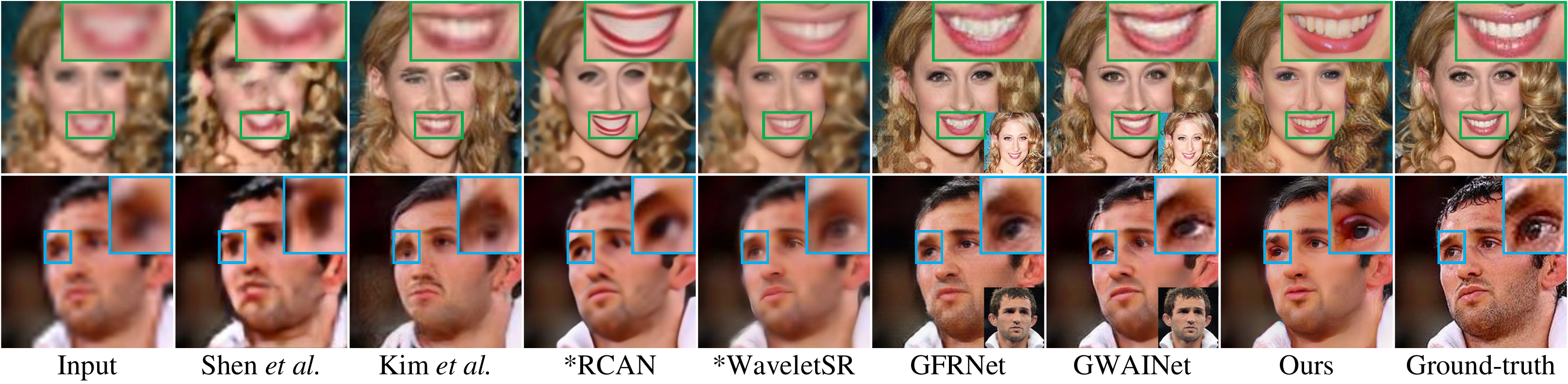}
	\caption{{{Visual comparisons of these competing methods on $\times8$ SR task. Close-up in the right bottom of GFRNet is the required guidance.}}}
	\label{fig:x8}
\end{figure*}

\subsubsection{Visual Comparisons.} Figs.~\ref{fig:x4} and \ref{fig:x8} show the restoration results of these competing methods on $\times4$ and $\times8$ super-resolution tasks. Shen \etal~\cite{shen2018deep} and Kim \etal~\cite{progressive_face_sr} were proposed to handle face deblur and super-resolution problems. Since they only released their test model, we did not re-implement them with the same training data and degradation model in this paper, resulting in their poor performance. The retrained *RCAN, *ESRGAN and *WaveletSR still limited in generating plausible facial structure, which may be caused by the lack of reasonable guidance for face restoration. In terms of reference-based methods, GFRNet~\cite{li2018learning} and GWAINet~\cite{dogan2019exemplar} generate plausible structures but fail to restore realistic details. In contrast to these competing methods, 
our DFDNet can reconstruct promising structure with richer details on these notable face regions (\ie, eyes and mouth). Moreover, even though the degraded input is not frontal, our DFDNet can also have plausible performance (2-\textit{nd} rows in Figs.~\ref{fig:x4} and \ref{fig:x8}).

\begin{figure*}[t]
	\setlength{\belowcaptionskip}{-16pt}
	\centering
	\includegraphics[width=1.\textwidth]{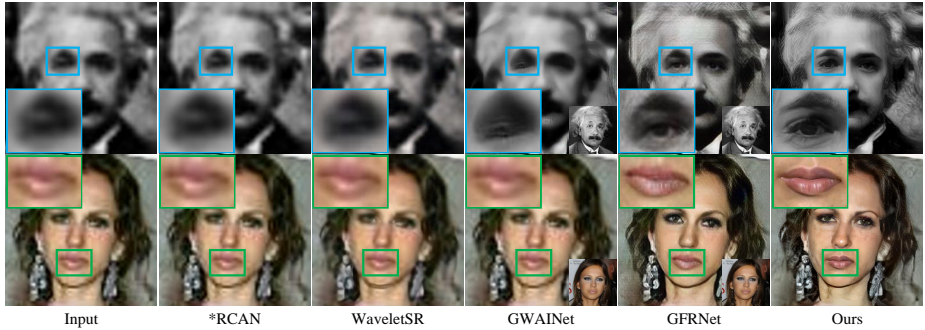}
	\caption{{Visual comparisons of competing methods with top performance on real-world low-quality images. Close-up at the right bottom is the required guidance. 
	}}
	\label{fig:real}
\end{figure*}

\subsubsection{Performance on Real-world Low-quality Images.} Our goal is to restore the real low-quality images without knowing the degradation types and parameters. To evaluate the performance of our DFDNet on blind face restoration, we select the real images from Google Image with face resolution lower than $80 \times 80$ and each of them has an identity-belonging high-quality reference for running GFRNet~\cite{li2018learning} and GWAINet~\cite{dogan2019exemplar}. Here we only show the visual results on competing methods with top-5 quantitative performance in Fig.~\ref{fig:real}. Among these competing methods, only GFRNet~\cite{li2018learning} is proposed to handle blind face restoration, thus can well generalize to real degraded images. However, its results still contain obvious artifacts due to the inconsistent reference of only one high-quality image. With the incorporation of component dictionaries, our DFDNet can generate plausible and realistic results, especially in the eyes and mouth region, indicating the effectiveness of our DFDNet in handling real degraded observations. Moreover, our DFDNet does not require the identity-belonging reference, showing practical values in wide applications.


\subsection{Ablation Study}
\begin{figure*}[t]
	\setlength{\abovecaptionskip}{3pt}
	\setlength{\belowcaptionskip}{-10pt}
	\centering
	\includegraphics[width=1.\textwidth]{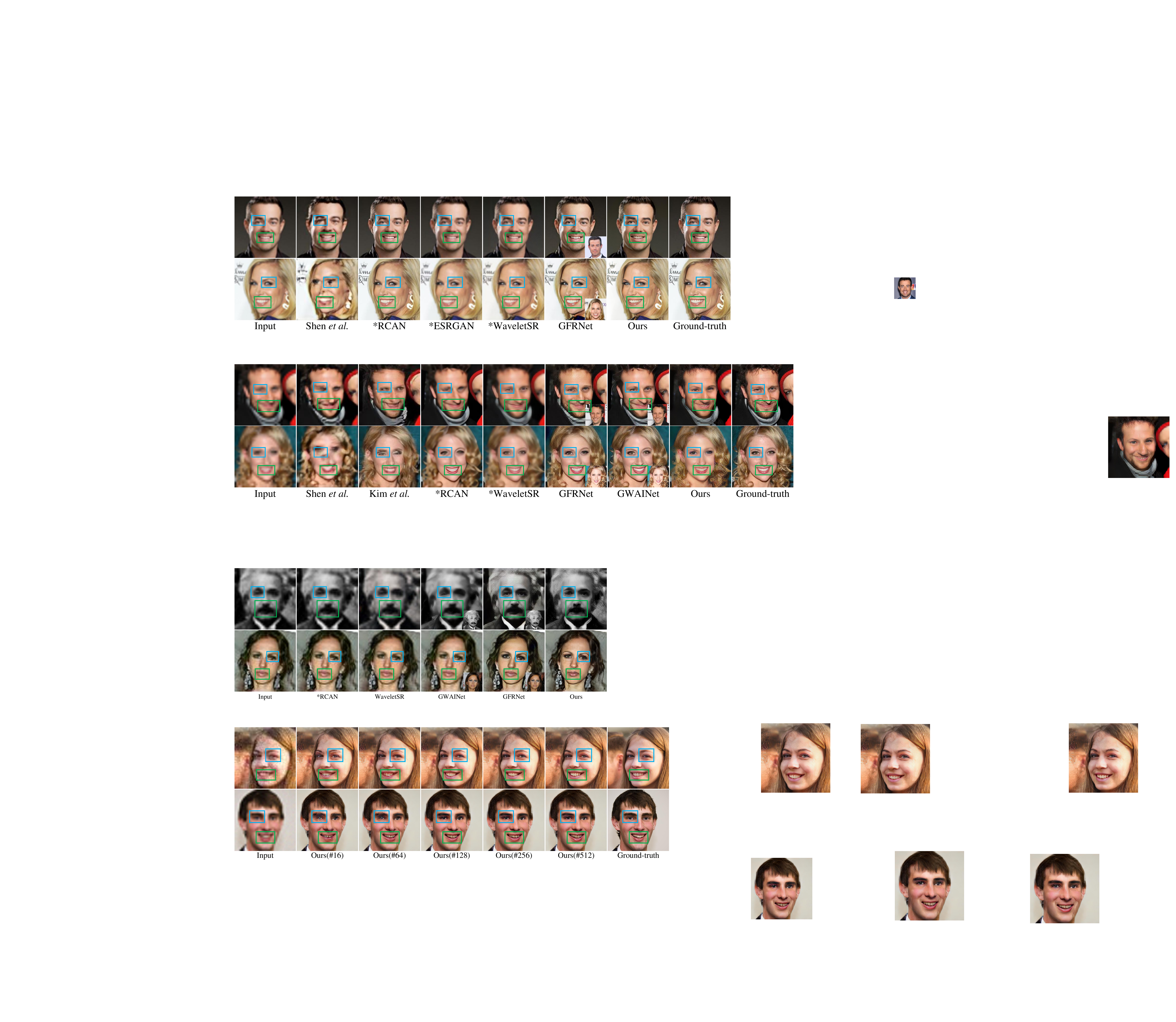}
	\caption{Restoration results of our DFDNet with different cluster numbers.}
	\label{fig:cluster}
\end{figure*}

To evaluate the effectiveness of our proposed DFDNet, we conduct two groups of ablative experiments, \ie, the cluster number $K$ for each component dictionary, and the progressive dictionary feature transfer block (DFT). For the first one, we generate different number of clusters in our component dictionaries. In this paper, we consider the cluster $K \in \{16,64,128,256,512\}$. For each variant, we retrain our DFDNet256 with the same experimental settings but with different cluster numbers, which are defined as Ours(\#$K$). The quantitative results on our VggFace2 test data are shown in Table~\ref{tab:cluster}. One can see that Ours(\#64) has nearly the same performance with GFRNet~\cite{li2018learning}. We analyze that because GFRNet~\cite{li2018learning} adopts alignment between reference and degraded input, making Ours(\#16) performs poorer than it. By increasing the cluster numbers, our DFDNet tends to achieve better results. We note that Ours(\#256) performs on par with Ours(\#512) but has less time-consuming in feature match. Thus, we adopt Ours(\#256) as our default model. Visual comparisons between these five variants are also presented in Fig.~\ref{fig:cluster}. 
We can see that when $K$ is larger, the restoration results tend to be clear and are much more realistic, indicating the effectiveness of our dictionaries in guiding the restoration process.

\begin{figure*}[t]
	\setlength{\abovecaptionskip}{3pt}
	\setlength{\belowcaptionskip}{-16pt}
	\centering
	\includegraphics[width=1.\textwidth]{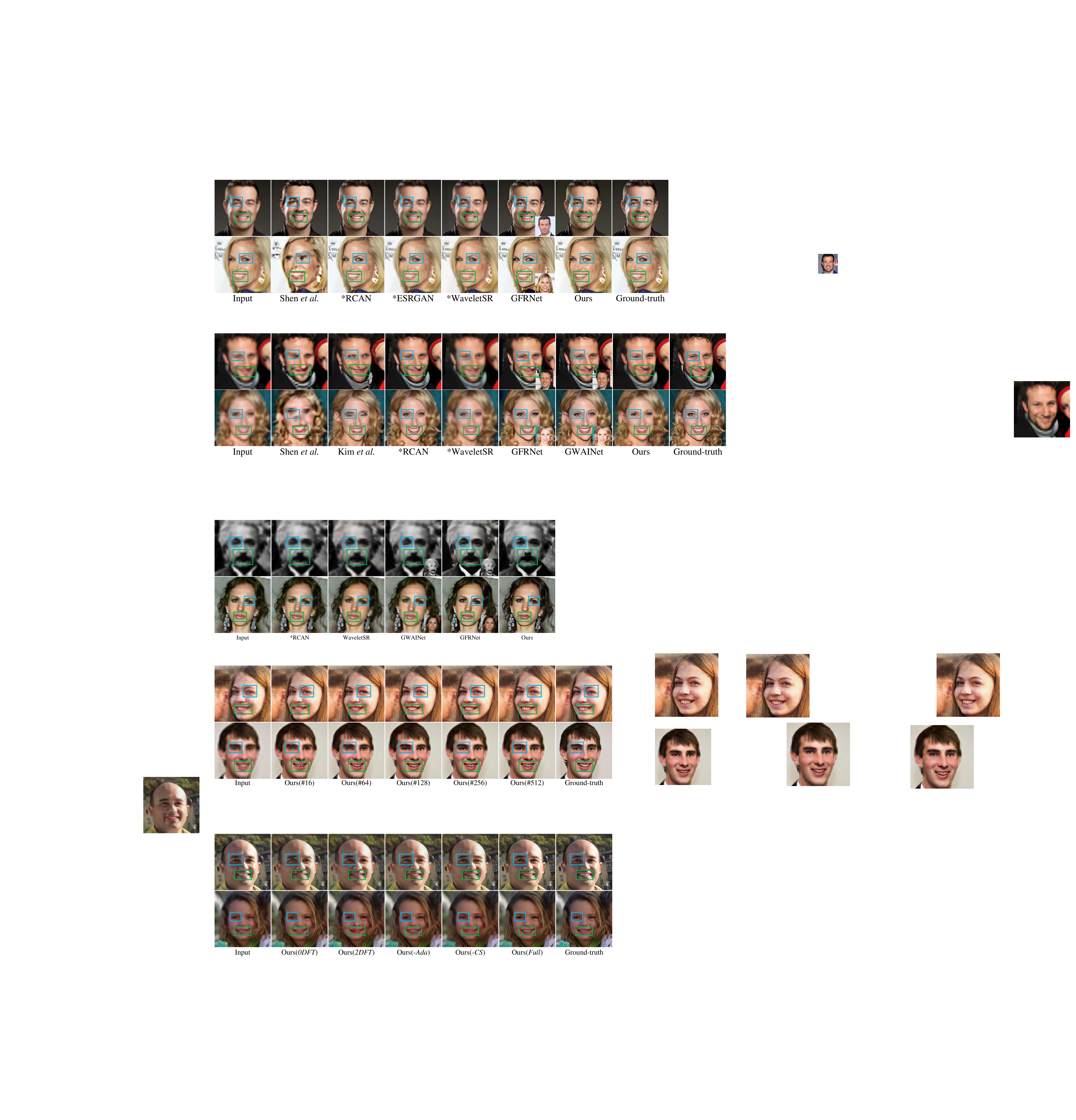}
	\caption{Restoration results of our DFDNet variants.}
	\label{fig:variant}
\end{figure*}
\begin{table}[t]
	\begin{minipage}{0.495\linewidth}
		\centering
		\caption{Comparisons on cluster number. }
		\label{tab:cluster}
		\scriptsize
		\setlength{\tabcolsep}{0.0mm}
		{
			\begin{tabular}{|c| c c c | c c c |} 
				\hline
				\rowcolor{lightgray}& \multicolumn{3}{c|}{$\times 4$} & \multicolumn{3}{c|}{$\times 8$}\\
				\rowcolor{lightgray}\multirow{-2}{*}{\makecell[c]{Methods}}&
				\tiny PSNR$\uparrow$ &\tiny SSIM$\uparrow$ & \tiny LPIPS$\downarrow$ & \tiny PSNR$\uparrow$ & \tiny SSIM$\uparrow$ & \tiny LPIPS$\downarrow$\\
				\hline \hline
				Ours(\#16) & 26.79 & .908 & .144 & .23.21 & .839 & .257 \\
				Ours(\#64) & 27.15 & .914 & .126  & 23.38 & .856 & .266 \\
				Ours(\#128) & 27.43 & .919 & .120  & 23.56 & .867 & .248 \\
				Ours(\#256) & 27.54 & .923 & .114  & 23.73 & .872 & .239 \\
				Ours(\#512) & 27.55 & .923 & .110  & 23.75 & .873 & .231 \\
				\hline
			\end{tabular} 
		}
	\end{minipage}
	\begin{minipage}{0.49\linewidth}  
		\centering
		\caption{Comparisons on variants of DFT. }
		\label{tab:var}
		\scriptsize
		\setlength{\tabcolsep}{0.0mm}
		{
			\begin{tabular}{|c| c c c | c c c |} 
				\hline
				\rowcolor{lightgray}& \multicolumn{3}{c|}{$\times 4$} & \multicolumn{3}{c|}{$\times 8$}\\
				\rowcolor{lightgray}\multirow{-2}{*}{\makecell[c]{Methods}}&
				\tiny PSNR$\uparrow$ &\tiny SSIM$\uparrow$ & \tiny LPIPS$\downarrow$ & \tiny PSNR$\uparrow$ & \tiny SSIM$\uparrow$ & \tiny LPIPS$\downarrow$\\
				\hline \hline
				Ours(\textit{0DFT}) & 25.30 & .896 & .239 & 23.06 & .839 & .253 \\
				Ours(\textit{2DFT}) & 26.43 & .905 & .161 & 23.24 & .848 & .261 \\
				Ours(\textit{-Ada}) & 25.47 & .897 & .190 & 22.97 & .836 & .270 \\
				Ours(\textit{-CS})  & 27.23 & .914 & .129 & 23.51 & .862 & .246 \\
				Ours(\textit{Full}) & 27.54 & .923 & .114 & 23.73 & .872 & .239 \\
				\hline
			\end{tabular} 
		}
	\end{minipage}
\end{table}

For the second one, to evaluate the effectiveness of our progressive DFT block, we consider the following variants: 
1) Ours(\textit{Full}): the final model in this paper, 
2) Ours(\textit{0DFT}): our DFDNet by removing all the DFT blocks and directly using SFT to transfer the encoder feature to the decoder, 
3) Ours(\textit{2DFT}): our DFDNet with two DFT blocks (\ie, DFT-\{3,4\} block), 
4) Ours(\textit{-Ada}) and Ours(\textit{-CS}): by removing the CAdaIN and Confidence Score in all the DFT blocks of final model, respectively. The quantitative results on our VggFace2~\cite{cao2018vggface2} test data are reported in Table~\ref{tab:var}. We can have the following observations. (i) By increasing the number of DFT block, obvious gains (at least 2.2 dB in $\times4$ and 0.6 dB in $\times8$) are achieved, indicating the effectiveness of our progressive manner. (ii) The performance is severely degraded when removing the CAdaIN. This may be caused by the inconsistent distribution of degraded feature and dictionaries, resulting in the wrong matched features for restoration. (iii) With the incorporation of confidence score, which can help balance the input and the matched dictionary feature, our DFDNet can also achieve plausible improvements. Fig.~\ref{fig:variant} shows the restoration results of these variants. We can see that compared with Ours(\textit{0DFT}) and Ours(\textit{2DFT}), Ours(\textit{Full}) is much clear and contains rich details. Results of Ours(\textit{-Ada}) are inconsistent with ground-truth (\ie, mouth region in 1-\textit{st} row). 
By the way, when the degradation is slight (1-\textit{st} row),  Ours(\textit{-CS}) which directly swaps the dictionary feature to the degraded image can easily change the original content (mouth region), making the undesired modification of face components.
\section{Conclusion}\label{section5}
In this paper, we present a blind face restoration model, \ie, DFDNet, to solve the limitation of reference-based methods. To eliminate the dependence of identity-belonging high-quality reference, we firstly suggest traditional K-means on large amount of high-quality images to cluster perceptually significant facial component. For dictionary feature transfer, we then propose a DFT block by addressing the following problems, distribution diversity between degraded input and dictionary feature with proposed component AdaIN, feature match scheme with fast inner product similarity, and generalization to degradation level with the confidence score. Finally, the multi-scale component dictionaries are incorporated in the multiple DFT blocks in a progressive manner, which can make our DFDNet learn the coarse-to-fine details for face restoration. Experiments validate the effectiveness of our DFDNet in handling the synthetic and real-world low-quality images. Moreover, we did not require an identity-belonging reference, showing the practical value in wide scenes in the real-world applications.

\subsubsection{Acknowledgments.}
This work is partially supported by the National Natural Science Foundation of China (NSFC) under Grant No.s 61671182, U19A2073 and Hong Kong RGC RIF grant (R5001-18).
\clearpage
%
%
\bibliographystyle{splncs04}
\bibliography{egbib}
\end{document}